\documentclass[letterpaper, 10 pt, conference]{ieeeconf}
\IEEEoverridecommandlockouts
\usepackage{amsmath,amssymb}
\usepackage{graphicx}
\usepackage{multicol}
\usepackage{footmisc}
\usepackage{float}
\usepackage{cite}
\usepackage[ruled,linesnumbered]{algorithm2e}
\usepackage[normalem]{ulem}
\usepackage{color}
\usepackage{subcaption}
\usepackage{url}
\usepackage{upgreek}
\usepackage{bm}

\newcommand{\hkg}[1]{{\textcolor{black}{#1}}}
\newcommand{\dg}[1]{{\textcolor{black}{#1}}}

\title{\LARGE \bf Event-Based Signal Temporal Logic Synthesis for Single and Multi-Robot Tasks}

\author{David Gundana and Hadas Kress-Gazit 
\thanks{D. Gundana and H. Kress-Gazit are with Sibley School of Mechanical and Aerospace Engineering, Cornell University, Ithaca, NY, 14853 USA. e-mail: \{dog4,hadaskg\}@cornell.edu. This work is supported by the National GEM Consortium, Cornell Sloan Fellowship, and NSF IIS-1830471. 
}%
}

\begin{document}

\maketitle
\thispagestyle{empty}
\pagestyle{empty}

\begin{abstract}

We propose a new specification language and control synthesis technique for single and multi-robot high-level tasks; these tasks include timing constraints and reaction to environmental events. Specifically, we define Event-based Signal Temporal Logic (STL) and use it to encode tasks that are reactive to uncontrolled environment events. Our control synthesis approach to Event-based STL tasks combines automata and control barrier functions to produce robot behaviors that satisfy the specification when possible. Our method automatically provides feedback to the user if an Event-based STL task can not be achieved. We demonstrate the effectiveness of the framework through simulations and physical demonstrations of multi-robot tasks. 
\end{abstract}

\section{INTRODUCTION}

High-level specifications have been used to describe complex robotics behaviors such as search and rescue missions and other planning and coordination tasks. Researchers have used control synthesis approaches to automatically generate controllers that satisfy high-level specifications described by temporal logic.  Temporal logics such as Linear Temporal Logic (LTL) are synthesized into controllers~\cite{Kress-Gazit2018} for single-robot systems, multi-robot systems (e.g. \cite{Loizou2004,Filippidis2012,Raman2014a}), and swarms (e.g. \cite{Kloetzer2007,Chen2018}). In other work, robot controllers have been synthesized for discrete-time continuous systems from Signal Temporal logic (STL) \cite{Donze2010} and Metric Temporal Logic (MTL) specifications \cite{Fainekos2009}. These specification languages can capture timing constraints associated with complex tasks \cite{Maler2004}. \hkg{In this paper we propose a specification formalism and associated control synthesis algorithm that combines the continuous (timing) properties of STL with the event-based nature of discrete logics, such as LTL, to enable users to specify tasks that have both timing constraints and desired reaction to external events.}

\hkg{\noindent\textbf{Synthesizing STL:}} Authors of \cite{Jones2019,Raman2014,Liu2017} present methods to design controllers for STL tasks. Work in \cite{Jones2019} provides a framework for solving a fragment of STL for multi-robot tasks. This method is robust to robot attrition and used for large teams of robots; however, the control is calculated before execution therefore it is not robust to disturbances encountered at runtime. The control synthesis approaches of \cite{Raman2014,Liu2017} provide robustness to disturbances. These methods rely on solving computationally expensive mixed-integer linear programs. The computation complexity makes it challenging to implement in real time, especially in the presence of dynamic obstacles.

\par
The authors of \cite{Ames2017} create control barrier functions (CBFs) and provide feedback control laws for a robot navigating in an environment with obstacles. These CBFs ensure that a system remains inside of a pre-defined set of allowable states, the safe-set, for all trajectories. \cite{Wang2017,Chen2018} leverage the work in \cite{Ames2017} to create safe control for multi-robot systems and swarms.

\par
The work in \cite{Lindemann2019b} uses time-varying CBFs to create a feedback control law that satisfies STL tasks for robotic systems in order to reduce the computational burden associated with solving mixed-integer linear programs.\cite{Lindemann2019a} extends \cite{Lindemann2019b} for multi-robot systems and introduces variables that relax CBFs and find a least violating solution when tasks conflict. Further, \cite{Lindemann2019c} creates a systematic procedure for constructing these CBFs to satisfy given STL tasks for multi-robot systems. In later work, \cite{Lindemann2019e} proposes a framework for satisfying STL tasks through automata based planning and timed signal transducers that represent temporal and Boolean operators \cite{Lindemann2019e}. We leverage \cite{Lindemann2019b,Lindemann2019a,Lindemann2019c, Lindemann2019e} in our work and extend its capabilities to include tasks that require the robot to react to events in the environment.
\par
\hkg{\noindent\textbf{Reactive STL:}} Researchers have investigated satisfying STL tasks that are reactive to external disturbances from the environment in order to encompass a larger set of complex tasks \cite{Raman2015}. These reactive STL tasks have been satisfied using model predictive control solved through mixed-integer linear programs. Disturbances are bounded and the authors make assumptions about the behaviour of the environment and adversaries in \cite{Raman2015}. In this paper, we propose a framework that considers these environment inputs to be discrete external events such as alarms and signals that have uncontrolled timings. To capture such tasks we create an extension of STL -- Event-based STL -- which can encode tasks where the robot must react to external events. 

\textbf{Assumptions:} In this paper, we assume that the initial state of the robot and the \hkg{initial state of the environment, that is whether environment events are triggered}\dg{,} do not violate the specification\dg{.} \hkg{In essence, we require that the system does not violate the specification before it starts executing.} \dg{We also assume that} all robots in the system are holonomic, \dg{meaning that the number of controllable degrees of freedom is equal to the total degrees of freedom.} \hkg{In multi-robot tasks we assume all robots can detect the state of the other robots, and that each robot computes its own control signal. }

\par
\textbf{Contributions: } We propose a framework for encoding tasks that contain timing constraints and reaction to environmental events, creating a control strategy to satisfy the task using CBFs, and providing feedback on the feasibility of these tasks. We present three main contributions: 1) a novel specification formalism, Event-based STL, that can capture \hkg{tasks that cannot be expressed in current STL synthesis techniques}, 2) an automata-based synthesis framework for generating decentralized controllers for multi-robot systems under an Event-based STL specification using time-varying CBFs, \hkg{thus reducing the computational burden of current reactive STL approaches} and 3) automated feedback to the user on the feasibility of Event-based STL tasks a-priori and at runtime for robots with bounded control inputs. 
\par


\section{Preliminaries}
\label{sec:prelim}

\subsection{Signal Temporal Logic (STL)}
\label{sec:STL}
Consider a \dg{continuous} time dynamical system representing robot motion:
\begin{equation}
    \dg{\dot{\textbf{x}} = f(\textbf{x}) + g(\textbf{x})\textbf{u}}
    \label{dynamics}
\end{equation}
Where \dg{\textbf{x}} $\in \mathbb{R}^n$ is the state of the system, \hkg{\textbf{u}}$\in \textbf{U} \subseteq \mathbb{R}^m$ is the bounded control input of the system, and $f$ and $g$ are locally Lipschitz continuous functions. 

Let $\mu \in \{True, False\}$ represent a predicate whose truth value is defined by the evaluation of a predicate function $h(\textbf{x}_t)$ \dg{where $\textbf{x}_t$ is the state of the system at time $t$}. 
\begin{equation}
    \mu ::= 
        \begin{cases}
            False & \Rightarrow h(\textbf{x}_t) < 0\\
            True & \Rightarrow h(\textbf{x}_t) \geq 0\\
        \end{cases}  
        \label{predicate}
\end{equation}

\noindent \textbf{Syntax}: An STL formula $\phi$ is defined recursively as 
\vspace*{-0.3mm}
\begin{equation}
    \phi ::= True\ |\ \mu\ |\ \neg \phi\ |\ \phi_1 \wedge \phi_2\ |\ F_{[a,b]}\phi\ |\ G_{[a,b]}\phi\ |\ \phi_1 U_{[a,b]}\phi_2
\end{equation}
where $\phi$ is an STL formula, a, b $\in \mathbb{R}^+$ are timing bounds, $\neg$ is ``not", $\wedge$ is ``and", $F$ is ``eventually", $G$ is ``always", and $U$ is ''Until" \cite{Maler2004}. 

\noindent \textbf{Semantics}: The semantics of STL are evaluated over trajectories of the dynamical system in eqn. \ref{dynamics}: 
\vspace{-.5mm}
\begin{table}[h]
\normalsize
\centering
\begin{tabular}{p{2.5cm} p{5.5cm}}
    $\textbf{x}_t \vDash \mu$ & $\Leftrightarrow h(\textbf{x}_t) \geq 0 \  $\\
    $\textbf{x}_t \vDash \neg \phi $ & $\Leftrightarrow \textbf{x}_t \not\vDash \phi \  $\\
    $\textbf{x}_t  \vDash  \phi_1 \wedge \phi_2$ & $\Leftrightarrow \textbf{x}_t \vDash \phi_1$ and $\textbf{x}_t\vDash \phi_2$\\
    $\textbf{x}_t \vDash F_{[a,b]} \phi$ & $\Leftrightarrow \exists t_1 \in [t+a,t+b] \ s.t. \  \textbf{x}_{t_1} \vDash \phi$\\
    $\textbf{x}_t \vDash G_{[a,b]} \phi$ & $\Leftrightarrow \forall t_1 \in [t+a,t+b], \  \textbf{x}_{t_1} \vDash \phi$\\
    
    \begin{tabular}{@{}c@{}}$\textbf{x}_t\vDash \phi_1 U_{[a,b]} \phi_2$ \\ \ \end{tabular} & \begin{tabular}{@{}c@{}}$\Leftrightarrow \exists t_2 \in [t+a,t+b] \  s.t.\  \textbf{x}_{t_2} \vDash \phi_2$ \\ and $\forall t_{1} \in [t+a,t_2], \textbf{x}_{t_1} \vDash \phi_1$ \end{tabular} \\

\end{tabular}
\end{table}

Intuitively, $F_{[a,b]}\phi$ is $True$ if there exists a time between $a$ and 
$b$ where $\phi$ is $True$, $G_{[a,b]}\phi$ is $True$ if $\phi$ is $True$ for all time between $a$ and $b$, and $\phi_1U_{[a,b]}\phi_2$ is $True$ if $\phi_1$ is $True$ \hkg{from time $a$} until $\phi_2$ becomes $True$\hkg{, and $\phi_2$ becomes $True$ sometime in the interval [$a,b$] }. 

\subsection{CBFs for STL specifications}
CBFs were proposed in~\cite{Wieland2007} and used to define safe-sets for a system and ensure that the safe-set is forward invariant: if a system starts in the set it will always stay in that set. CBFs ensure forward invariance without determining the entire reachable set of system. Lindemann and Dimarogonas~\cite{Lindemann2019b} propose a process to generate control for robotic systems to satisfy STL formulas using CBFs. To ensure a task is satisfied given the timing constraints of an STL formula, \cite{Lindemann2019b} creates CBFs $cbf(\textbf{x}_t)$ that are time varying and forward invariant. These CBFs are constructed using predicate functions of the STL formula, $h(\textbf{x}_t)$. \hkg{Given a CBF, if eqn. \ref{cost1} holds for all $\textbf{x}_t$, then the}  \dg{system} is forward invariant, i.e. if $cbf(\textbf{x}_0) \geq 0$, then $cbf(\textbf{x}_t) \geq 0$ for all $t$.
\vspace*{-.2mm}
\begin{equation}
     \sup\limits_{\textbf{u} \in \textbf{U}}  \frac{\partial cbf(\textbf{x}_t)^T}{\partial x}(f(\textbf{x}_t) + g(\textbf{x}_t)\textbf{u}_t) 
     + \frac{\partial cbf(\textbf{x}_t)}{\partial t} \geq -\nu(cbf(\textbf{x}_t))
     \label{cost1}
\end{equation}
where $\nu: \mathbb{R}_{\geq 0} \rightarrow \mathbb{R}_{\geq 0}$ is a locally Lipschitz continuous \dg{class $K$} function. 

\hkg{In~\cite{Lindemann2019b} III.A, the authors describe in detail the process for generating CBFs such that if we can find a control input $u$ that satisfies \dg{eqn.} \ref{cost1}, the system is guaranteed to satisfy the associated STL formulas with timing constraints. Intuitively, we require (i) the CBF to be valid, i.e. satisfy $cbf(\textbf{x}_t) \geq 0, \forall t$. To ensure the timing constraints in the STL formulas are met, we require (ii) $cbf(\textbf{x}_t) \leq h(\textbf{x}_t)$ based on predicate $\mu$ and the associated temporal operator; at some $t$ in the interval for $F_{[a,b]}$, for all $t$ in the interval for $G_{[a,b]}$, etc. The combination of these constraints guarantees $h(\textbf{x}_t)\geq 0$ at the required times thus satisfying the STL formula.}

 We can combine STL formulas and create controllers that do not violate any of the individual CBFs in order to express more complex tasks. This is done using an approximation for the minimum of the barrier functions for each task. One can design a single CBF, $cbf_{\phi}$, such that if $cbf_{\phi} \geq 0$, then  $cbf_{i} \geq 0$  $\forall i$ \cite{Lindemann2019b}:
\vspace*{-.3mm}
\begin{equation}
    cbf_{\phi} = -\ln\Big(\sum_{i=1}^{I} exp(-cbf_i(\textbf{x}_t))\Big)
    \label{combine}
\end{equation}

where $I$ is the number of CBFs in a given specification.
\subsection{Linear Temporal Logic (LTL) and B\"{u}chi Automata}
An LTL formula $\gamma$ is constructed from a set of atomic propositions $AP$ using the following grammar
\vspace{-.5mm}
\begin{equation}
    \gamma ::= \pi |\  \neg \gamma \ |\  \gamma_1 \wedge \gamma_2 \ |\  X \gamma\ |\ \gamma_1 U \gamma_2
\end{equation}
where $\pi \in AP$, $\neg$ and $\wedge$ are the Boolean operators ``not" and ``and", $X$ is the temporal operator ``next", and $U$ is the temporal operator ``Until". From these operators we can define the temporal operators ``eventually" ($F\gamma = True U \gamma$) and ``always" ($G\gamma = \neg F\neg\gamma$). The semantics of LTL are defined over an infinite sequence $\sigma = \sigma_1, \sigma_2 ...$, where $\sigma_i \subseteq AP$ represents the \hkg{set of} propositions that are $True$ in position $i$ of the sequence. 



Intuitively, $X\gamma$ is $True$ if for every execution $\gamma$ is $True$ in the next position of the sequence, $F\gamma$ is $True$ if for every execution $\gamma$ is $True$ at some position in the sequence, $G\gamma$ is $True$ if for every execution $\gamma$ is $True$ at all positions of the sequence, and $\gamma_1U\gamma_2$ is $True$ if for every execution $\gamma_1$ is $True$ until $\gamma_2$ becomes $True$.

\hkg{A nondeterministic} B\"{u}chi automaton is a tuple
\begin{equation}
    B = (S,s_0,\Sigma,\delta,F)
    \label{buchi}
\end{equation}
where $S$ is a finite set of states, $s_0$ is the initial state, $\Sigma$ is a finite input alphabet, $\delta$ $\subseteq S \times \Sigma \times S$ is the transition relation, and $F\subseteq S$ is a set of accepting states. A run of a B\"{u}chi automaton on input word $\omega=\omega_1,\omega_2..., \; \omega_j\in\Sigma$ is an infinite sequence of states $s_0,s_1,s_2,...$ s.t. $\forall j\geq 1, \hkg{(s_{j-1},\omega_j,s_{j})\in\delta}$. We define $\inf(\omega)$ as a set of states that are visited infinitely often on the input word $\omega$. A run is accepting iff $\inf(\omega) \cap F \neq 0$. 
\par
Given an LTL formula $\gamma$, one can construct a \hkg{nondeterministic} B\"{u}chi automaton $B_\gamma$ such that $B_\gamma$ only accepts input words that satisfy $\gamma$~\cite{Gastin2001,Holzmann1997}. In the following we use the LTL to B\"{u}chi automaton tool Spot \cite{Duret-Lutz2016}.



\section{Event-Based STL}
\label{sec:EvSTL}
 We define a new specification formalism, Event-based STL, to describe tasks that have not been previously addressed by STL synthesis techniques. This formalism can capture tasks where the system needs to react to uncontrolled environmental events that may or may not occur during execution. Examples of these events are fire alarms in an evacuation scenario, a person entering in a room in a workspace environment, or a command from a user. 

\subsection{System Representation}
The system model is defined by eqn. \ref{dynamics}. In addition to the system model, we consider discrete environmental events. These environmental events are uncontrolled by the system and are represented as Boolean propositions $\pi \in AP$. We define $\sigma_t$ $\subseteq$ $AP$ as the set of atomic propositions that are $True$ at time $t$.

\subsection{Syntax of Event-Based STL}
We define Event-based STL formulas $\Psi$ as follows:
\begin{align}
    \varphi ::=  &\mu \ |\ \neg \mu\ | \   \varphi_1 \wedge \varphi_2\\
    \alpha ::= &\pi \  | \ \neg \alpha \ |\   \alpha_1 \wedge \alpha_2 \  \\
    \begin{split}
    \Psi ::= &G_{[a, b]}\ \varphi \ |\ F_{[a, b]}\ \varphi \ | \ \varphi_1 \ U_{[a,b]} \ \varphi_2 \ |\\ &\ G(\alpha \Rightarrow \ \Psi) \ | \ G(\varphi \Rightarrow \ \Psi) \ | \ \Psi_1 \wedge \Psi_2
    \end{split}
\end{align}
where $\mu$ is a predicate representing $h(\textbf{x}_t)$  as described in eqn. \ref{predicate}, $\alpha$ is a Boolean formula over environment propositions $\pi \in AP$, $\Rightarrow$ is the implication operator, and the temporal operators follow the conventions of STL, as defined in Sec. \ref{sec:STL}. If the "always" operator $G$ does not contain a timing bound $[a,b]$, we assume the timing bound is $[0, \infty]$. \dg{We \hkg{interpret} the interval $[a,b]$ in continuous time similar to the fragment of STL in \cite{Lindemann2019b}. \hkg{When executing the control on a simulated or physical system with a clock,} we evaluate the dynamics of the system and the environment at a set sampling rate. For demonstrations we assume that the interval $[a,b]$ is a multiple of this sampling rate} \hkg{and environment events last longer than the sampling period.}

\subsection{Semantics of Event-Based STL}
We define the semantics of Event-based STL over $(\textbf{x}_t,\sigma_t)$ where $\textbf{x}_t$ is the state of the system at time $t$ and $\sigma_t$ is a set of environment propositions that are $True$ at time $t$. 

\begin{table}[h]
\normalsize
\centering
\setlength\tabcolsep{0.01 pt}
\begin{tabular*}{\textwidth}{p{3.145cm} p{5.48cm}}
    $(\textbf{x}_t,\sigma_t) \vDash \mu$ & $\Leftrightarrow h(\textbf{x}_t)$ $ \geq 0 \  $\\
    $(\textbf{x}_t,\sigma_t) \vDash \neg\mu $
    & $\Leftrightarrow h(\textbf{x}_t)<0 \  $\\
    $(\textbf{x}_t,\sigma_t)  \vDash  \varphi_1 \wedge \varphi_2$ & $\Leftrightarrow (\textbf{x}_t,\sigma_t) \vDash \varphi_1$ and $(\textbf{x}_t,\sigma_t)\vDash \varphi_2$\\
    \\

    $(\textbf{x}_t,\sigma_t) \vDash \pi$ & $\Leftrightarrow \pi \in \sigma_t$  \\
    $(\textbf{x}_t,\sigma_t) \vDash \neg \alpha$ & $\Leftrightarrow (\textbf{x}_t,\sigma_t) \nvDash \alpha$\\
    $(\textbf{x}_t,\sigma_t) \vDash \alpha_1 \wedge \alpha_2$ & $\Leftrightarrow (\textbf{x}_t,\sigma_t) \vDash \alpha_1$ and $ (\textbf{x}_t,\sigma_t) \vDash \alpha_2$\\

    \\
    $(\textbf{x}_t,\sigma_t) \vDash F_{[a,b]} \varphi$ & $\Leftrightarrow \exists t_1 \in [t+a,t+b] \ s.t. \  (\textbf{x}_{t_1},\sigma_{t_1}) \vDash \varphi$ \\
    $(\textbf{x}_t,\sigma_t) \vDash G_{[a,b]} \varphi$ & $\Leftrightarrow \forall t_1 \in [t+a,t+b], \ (\textbf{x}_{t_1},\sigma_{t_1}) \vDash \varphi$ \\
    
   $(\textbf{x}_t,\sigma_t)\vDash \varphi_1 U_{[a,b]} \varphi_2$ & $\Leftrightarrow \exists t_2 \in [t+a,t+b] s.t. \ (\textbf{x}_{t_2},\sigma_{t_2}) \vDash \varphi_2$ and $\forall t_{1} \in [t+a,t_2],(\textbf{x}_{t_1},\sigma_{t_1}) \vDash \varphi_1 $\\
    
    $(\textbf{x}_t,\sigma_t) \vDash G(\alpha \Rightarrow \Psi) \ $ & $\Leftrightarrow  \forall t, (\textbf{x}_t,\sigma_t) \nvDash \alpha $ or $ (\textbf{x}_t,\sigma_t) \vDash \Psi $ \\
    $(\textbf{x}_t,\sigma_t) \vDash G(\varphi \Rightarrow \Psi)\ $ & $\Leftrightarrow \forall t, (\textbf{x}_t,\sigma_t) \nvDash \varphi$ or $(\textbf{x}_t,\sigma_t) \vDash \Psi$\\ 
    $(\textbf{x}_t,\sigma_t)  \vDash  \Psi_1 \wedge \Psi_2$ & $\Leftrightarrow (\textbf{x}_t,\sigma_t) \vDash \Psi_1$ and $(\textbf{x}_t,\sigma_t) \vDash \Psi_2$\\
    
\end{tabular*}
\end{table}


\section{Problem Formulation}
\label{sec:problem}
\textbf{Problem:} Given a dynamical system (eqn. \ref{dynamics}) and its state $\textbf{x}$, environment events $AP$, and an Event-based STL formula $\Psi$, find control $u$ such that $(\textbf{x}_0,\sigma_0)  \vDash  \Psi$ , if possible.

We describe our approach to synthesizing the control in Section \ref{sec:synth}, and discuss feedback and guarantees in Section \ref{sec:feedback}. For multi-robot tasks, we propose a decentralized control strategy \dg{where each robot satisfies the Event-based STL \hkg{formulas} that affect them. \hkg{We do this} in order to distribute the computational burden associated with finding control inputs for each robot. In this framework} each robot \dg{knows} the position of the other robots \dg{in the system}, but not their control inputs \hkg{or goals}.

\subsection{Examples}
\textbf{Single-Robot Example: } We consider a holonomic robot operating in an obstacle-free workspace. The robot's motion is described by eqn. \ref{dynamics} where $\textbf{x}_t \in \mathbb{R}^2$ is the state of the robot $[x_t,y_t]^T$ at time $t$, \dg{$f(\textbf{x}_t) = 0$}, $g(\textbf{x}_t) = I_2$, and $\textbf{u}_t$ is the control input $[u_{x_t},u_{y_t}]^T$. We define $AP=\{alarm\}$ as the set of environment events. The robot's task is whenever it senses the alarm, to arrive, within 10 time steps, at a point within 1 unit from $[5,5]$. The task is captured by the Event-based STL formula $$\Psi = G(alarm \Rightarrow F_{[0,10]}(\parallel \textbf{x} -[5,5]^T \parallel < 1))$$ Here, $h(\textbf{x}_t)=(1- \parallel \textbf{x}_t -[5,5]^T \parallel )$.

\textbf{Multi-Robot Example: }We consider four holonomic robots operating in an obstacle-free workspace. The dynamics of the robots are described by eqn. \ref{dynamics}, where $\textbf{x}$ describes the state of the robots $\textbf{x} = [\textbf{x}_1, \textbf{x}_2, \textbf{x}_3, \textbf{x}_4]$, $\textbf{x}_i = [x_{i,t},y_{i,t},\theta_{i,t}]$\dg{, $f(\textbf{x}_i) = 0$, $g(\textbf{x}_i) = $ \hkg{$I_3$}, $\textbf{u} = [\textbf{u}_1,\textbf{u}_2,\textbf{u}_3,\textbf{u}_4]$, and $\textbf{u}_i = [u_{i,x_t}, u_{i,y_t}, u_{i,\theta_t}]$} for each robot $i$.

 We define $AP = \{approach, align\}$. The multi-robot task is captured by the following Event-based STL formula $\Psi = \Psi_1 \wedge \Psi_2 \wedge \Psi_3 \wedge \Psi_4 \wedge \Psi_{collision} \wedge \Psi_{approach} \wedge \Psi_{align}$ where the subformulas are 
\begin{itemize}
    \item $\Psi_1 = F_{[0,10]}(\parallel \textbf{x}_1 - [3,1]^T \parallel < 0.5) $
    \item $\Psi_2 = F_{[5,15]}(\parallel \textbf{x}_2 - [3,2]^T \parallel < 0.5) $
    \item $\Psi_3 = F_{[0,10]}(\parallel \textbf{x}_3 - [3,0]^T \parallel < 0.5) $
    \item $\Psi_4 = F_{[0,10]}(\parallel \textbf{x}_4 - [3,2]^T \parallel < 0.5) $
    
    \item $\Psi_{collision_{ij}} = G_{[0,30]}(\parallel \textbf{x}_i - \textbf{x}_j \parallel > 0.3), \   \forall i \neq j $
    
   \item $\Psi_{approach_i} = G(approach \Rightarrow F_{[0,10]} (\parallel\textbf{x}_i - [6,2]^T \parallel < 1)), \   i = 1,3$
   
   \item $\Psi_{align_i} = G(align \Rightarrow F_{[0,10]} (\lvert\, \lvert\theta_i\rvert - \pi \ \rvert < 0.1)), \   i = 2,4$

\end{itemize}
Subformulas $\Psi_{1,2,3,4}$ describe when the robots should be in a certain region. $\Psi_{collision_{ij}}$ describes six subformulas for collision avoidance which states that each robot must maintain a distance of at least 0.3 units from every other robot. $\Psi_{approach_i}$ states that, for robots 1 and 3, if the environment event $approach$ is sensed, then they should arrive close to $[6,2]$ (no more than 1 away) within 10 time units. $\Psi_{align_i}$ states that, for robots 2 and 4, if the environment event $align$ is sensed, they both should, within 10 time units, be facing the -x direction of the global reference frame.

\section{Synthesis for Event-based STL}
\label{sec:synth}

Algo. \ref{algo:synth} describes our approach to automatically synthesizing control given a high-level task encoded in Event-based STL. The inputs to this algorithm are an Event-based STL formula $\Psi_{STL}$, the number of robots $n$, $\sigma_t$, $\textbf{x}_t$, and the functions $h_i(\textbf{x}_t)$. The outputs are the control inputs $\textbf{u}_i \in \textbf{U}_i$ for each robot, that satisfy $\Psi_{STL}$. 

Algo. \ref{algo:synth} has two phases; first, before execution, we create template CBFs based on the predicates in $\Psi$ and create a B\"{u}chi automaton that we use to temporally compose CBFs based on environmental events (Section \ref{subsec:beforeExecution}). Then, during execution, we choose a transition in the B\"{u}chi automaton that corresponds to the currently sensed events in the environment (Section \ref{subsec:transition}) and create the control from the CBFs that correspond to that transition (Section \ref{subsec:control}.) \dg{The B\"{u}chi automaton \hkg{acts as} a centralized \hkg{symbolic} planner. \hkg{Given a transition in the B\"{u}chi automaton,} each robot determines their own control strategy in a decentralized manner.}

\begin{algorithm}[h]
\SetAlgoLined
\SetKwInOut{Input}{Input}
\SetKwInOut{Output}{Output}
\SetKwProg{Initialization}{Initialization}{}{}

\Input{$\Psi_{STL}$, $n$, $\sigma_t$, $\textbf{x}_t$, $h_i(\textbf{x}_t),$}
\Output{$\textbf{u}$}
$\forall i, cbf_{\mu_i} = CBFTemplate(h_i(\textbf{x}_t))$\; \label{line:bfunc}
$(\Psi_{LTL},\Pi_{\mu}) = STL2LTL(\Psi_{STL})$\; \label{line:stl2ltl}
$B_{\Psi_{LTL}} = LTL2Buchi(\Psi_{LTL})$\;\label{line:ltl2B}
$currS = s_0$\;
$\sigma_{-1} = \sigma_0$\;
$(\bm{\upsigma}_{currS,nextS},\Pi_{\mu_{act}},currS) = findTransition(\sigma_0,\textbf{x}_0,B_{\Psi_{LTL}},h_i(\textbf{x}_0),currS)$\;
\label{line:findTransition}
\While{True}{
    \tcp{check whether reached $nextS$ or environment event changed}
    \If{($\bm{\upsigma}_{currS,nextS}$ is $True$) or $\sigma_t \neq \sigma_{t-1}$ \label{line:nextState} }{
    $(\bm{\upsigma}_{currS,nextS},\Pi_{\mu_{act}},currS) = findTransition(\sigma_t,\textbf{x}_t,B_{\Psi_{LTL}},h_i(\textbf{x}_t),currS)$\;
    }
   \tcp{Execute Barrier Functions}
   \For{i = 1 to n}{
   $\textbf{u}_i = Barrier(\Pi_{\mu_{act}},t,\textbf{x}_t)$\;\label{line:commands}
   }
   \If{Eqn. \ref{cost} is infeasible}{
   Stop\;
   }
   $\sigma_{t-1} = \sigma_t$\; }
\caption{Control synthesis for Event-based STL}
\label{algo:synth}

\end{algorithm}

\subsection{CBFs and Abstracted Automaton}
\label{subsec:beforeExecution}

Given an Event-based STL formula $\Psi_{STL}$ we first create template CBFs corresponding to the predicates in $\Psi_{STL}$. We then abstract the formula into an LTL formula $\Psi_{LTL}$ and create a B\"{u}chi automaton $B_{\Psi_{LTL}}$ that we use to choose the CBFs that are executed. 

\textbf{Creating CBF templates $cbf_{\mu_i}$ (Line \ref{line:bfunc} of Algo. \ref{algo:synth}):} Given an STL formula $\phi$, Lindemann et. al. \cite{Lindemann2019b} provide a method for constructing CBFs that satisfy time constrained STL specifications assuming unbounded control. In this work we use the methods from \cite{Lindemann2019b} to create CBF formula templates that use parameters from an Event-based STL formula. 

We create a control barrier function template, eqn. \ref{bfunceq}, that changes linearly with time and utilizes the entirety of the time bound that is given. For this template we use the predicate function $h_i(\textbf{x}_t)$, the time that the CBF is initially activated $t_{int}$, and $a,b$ as place holders for the exact timing bounds specified in the subformulas of $\Psi_{STL}$; these timing bounds will be instantiated during execution (Sec. \ref{subsec:control}).
\vspace{-.5mm}
\begin{equation}
cbf_{\mu_i}(\textbf{x}_t ) = \frac{(t-t_{int}-a)h_i(\textbf{x}_{t_{int}})}{b-a} - h_i(\textbf{x}_{t_{int}}) + h_i(\textbf{x}_t)
\label{bfunceq}
\end{equation}

 We create CBFs in this way so that a robot has the greatest opportunity to satisfy its task. The CBF changing linearly and using the entire time bound represents a worst-case scenario of the safe-set at a point in time. At $t = t_{int} + a$, the initial time the CBF becomes activated,  $cbf_{\mu_i}(\textbf{x}_t ) = 0$. At $t = t_{int} + b$, the final time in the interval for the Event-based STL formula, $cbf_{\mu_i}(\textbf{x}_t ) = h_i(\textbf{x}_t)$.

For example, the predicate from the single robot example is $\mu_1=\parallel \textbf{x} – [5, 5]^T \parallel < 1 $. We form a predicate function $h_1(\textbf{x}) = 1 - \parallel \textbf{x} – [5, 5]^T \parallel $ and construct a \hkg{template} CBF $cbf_{\mu_1}(\textbf{x}_t) =  \frac{(t - t_{int}-a)(1 - \parallel \textbf{x}_{t_{int}} – [5, 5]^T \parallel)}{b-a} + \parallel\textbf{x}_{t_{int}} – [5, 5]^T\parallel - \parallel \textbf{x}_{t} – [5, 5]^T \parallel$.

\noindent \textbf{ Abstracting $\Psi_{STL}$ (Line \ref{line:stl2ltl} of Algo. \ref{algo:synth}):} We abstract $\Psi_{STL}$ to $\Psi_{LTL}$ by replacing $F_{[a,b]}$ with $F$, $G_{[a,b]}$ with $G$, and $U_{[a,b]}$ with $U$. Furthermore, we replace each $\mu_i$ with a proposition $\pi_{{\mu_i},[a,b]} \in \Pi_{\mu}$ that we consider a controllable proposition. Each controllable proposition maintains the timing associated with its Event-based STL subformula. For example, eqn. \ref{asSTL} is abstracted to eqn. \ref{asLTL}. 
\vspace{-.5mm}
\begin{gather} 
	\Psi_{STL} = G(alarm \Rightarrow F_{[0,10]}(\parallel \textbf{x} – [5, 5]^T \parallel < 1))
	\label{asSTL}\\
	\Psi_{LTL}  = G(alarm \Rightarrow F(\pi_{\mu_1,[0,10]})) 
	\label{asLTL}
\end{gather}
where $\pi_{\mu_1,[0,10]}$ replaces $(\parallel \textbf{x} – [5, 5]^T \parallel < 1)$ and the associated timing constrains. \hkg{Event-based STL allows limited nesting of temporal operators through the formulas with implications, as long as the antecedent is either a Boolean combination of environment propositions, or a conjunction of predicates and their negation.} \dg{For example, eqn. \ref{impl} is abstracted to eqn. \ref{implAb}}
\hspace*{-0.3mm}
\dg{
\begin{gather}
    \Psi_{STL} = G(\textbf{A} \Rightarrow G(\textbf{B} \Rightarrow F_{[0,10]}(\parallel \textbf{x} – [5, 5]^T \parallel < 1)))
    \label{impl} \\
    \Psi_{LTL} = G(\textbf{A} \Rightarrow G(\textbf{B} \Rightarrow F(\pi_{\mu_1,[0,10]})))
    \label{implAb}
\end{gather}}
\dg{Where $\textbf{A}$ and $\textbf{B}$ are external environment events. \hkg{We do} not allow nesting of any other form.}

\begin{figure}[h]
\centering
\includegraphics[width=.9\linewidth]{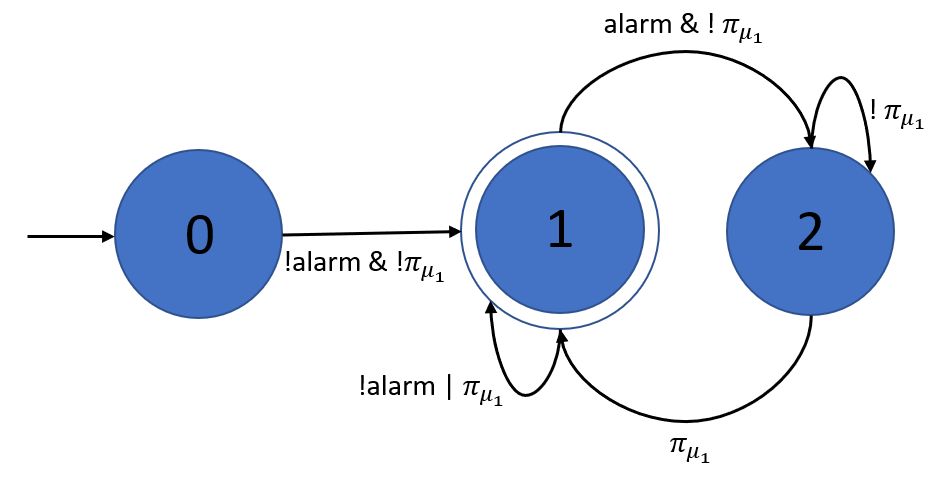}
\caption{Graphical representation of $B_{\Psi_{LTL}}$ for $\Psi_{LTL}$ in eqn. \ref{asLTL}. The grey circles represent states and the double circle represents an accepting state. Transitions between states are labeled with the Boolean formulas $\bm{\upsigma}_{s_i,s_j}$.}
\label{Buchi}
\end{figure}

\noindent \textbf{Generating $B_{\Psi_{LTL}}$ (Line \ref{line:ltl2B} of Algo. \ref{algo:synth}):} We create a B\"{u}chi automaton $B_{\Psi_{LTL}}$ from $\Psi_{LTL}$ using~\cite{Duret-Lutz2016}.  The transitions are labeled with Boolean formulas  over the set $AP \cup \Pi_\mu$ as seen in figure \ref{Buchi}. We denote a Boolean formula over $AP \cup \Pi_\mu$ representing the label of the transition between $s_i$ and $s_j$  as $  \bm{\upsigma}_{s_i,s_j}$, i.e. $\Sigma=\{\bm{\upsigma}_{s_i,s_j}\mid \exists s_i,s_j\in S, \; \hkg{(s_i,\bm{\upsigma}_{s_i,s_j},s_j)\in\delta}\}$ 

Here $S=\{s_0,s_1,s_2\}$, $s_0$ is the initial state, and $s_1$ is an accepting state. The transitions between states are labeled with Boolean formulas over $\{alarm,\pi_{\mu_1,[0,10]}\}$. For example, $\bm{\upsigma}_{s_0,s_1}=\neg alarm \wedge \neg\pi_{{\mu_1},[0,10]}$. The task is satisfied when the system is in the accepting state \dg{and can remain in a cycle that contains an accepting state}, \hkg{in this case} when the predicate $\mu_1$ is $True$ or the alarm is not activated. 
At runtime, based on the environment events and state of the system, we choose the next transition in the automaton, and then create the control to drive the robot(s).

\subsection{Choosing transitions}
\label{subsec:transition}

\noindent \textbf{Determining $\Pi_{\mu_{act}}$(Lines \ref{line:findTransition} and \ref{line:nextState} of Algo. \ref{algo:synth}):} 
During execution, we create the control for the robot(s) based on the label of the active transition in $B_{\Psi_{LTL}}$. The active transition is the transition the system is currently trying to take, by activating the CBFs associated with the controllable propositions $\Pi_\mu$.

At each time step, given $\sigma_t$, the set of environment propositions that are $True$, and the state of the system $\textbf{x}_t$, we first determine the truth values of all the propositions $AP \cup \Pi_\mu$; for $\pi\in AP$: 
\begin{equation}
    \pi = 
        \begin{cases}
            False & \text{if  } \pi\not\in \sigma_t \\
            True & \text{if  } \pi \in \sigma_t \\
        \end{cases}  
        \label{pi_mu}
\end{equation}
and for $\pi_{\mu_i,[a,b]}\in\Pi_\mu$:
\begin{equation}
    \pi_{\mu_i,[a,b]} = 
        \begin{cases}
            False & \text{if  } h_i(\textbf{x}_t) < 0\\
            True & \text{if  } h_i(\textbf{x}_t) \geq 0\\
        \end{cases}  
        \label{pi_mu}
\end{equation}


We then evaluate whether we need to find a new active transition; this would happen under two conditions, either (1) the environment propositions changed, i.e. $\sigma_t \neq \sigma_{t-1}$ which could change the truth value of the formula labeling the transition $\bm{\upsigma}_{currS,nextS}$, or (2) $\bm{\upsigma}_{currS,nextS}$ becomes $True$ indicating that all the associated predicates $\mu_i$ are $True$ and the system transitioned to the next state. 

If one of the above conditions holds, we choose a new active transition. To choose one, we first find the set of possible transitions. The system can choose to take transitions that are consistent with the current truth value of the (uncontrollable) environment propositions $AP$. Put another way, the set of possible transitions excludes transitions where the truth values of the propositions in $AP$ would cause $\bm{\upsigma}$ to evaluate to $False$.


\hkg{Given the set of possible transitions, we search for a path that will cause the LTL specification to be satisfied. Specifically, we find a $prefix$ -- a path to an accepting state and a $suffix$ -- a cycle that will cause the system to infinitely visit accepting states; the $suffix$ may be a self transition on an accepting state. If there are multiple paths, we choose the shortest prefix path; that being said, due to the structure of the formulas we allow (no disjunction over predicates or temporal formulas, no formulas of the form $FG$), and as is the case with all the examples in this paper, the B\"{u}chi automaton $B_{\Psi_{LTL}}$ is typically deterministic; therefore, the choice in paths is due to the possible different values of the controlled propositions. When deciding on the next state, if there are multiple paths of the same length, we choose the path where the next transition has the lowest number of $\pi_{\mu_i,[a,b]}$ that are $True$, thereby reducing the number of CBFs we need to activate.}

\hkg{Given the path, we choose as the active transition the next transition in this path. We denote the set of $\pi_{\mu_i,[a,b]}$ propositions that must be $True$ to satisfy the Boolean formula $\bm{\upsigma}_{currS,s_j}$ for this transition as $\Pi_{\mu_{act}}$; this set represents the CBFs that are activated for that transition to complete. }

Using the example, if the system is in state $s_2$, the only path to an accepting state is from state $s_2$ to state $s_1$. For this transition to occur, the controllable proposition $\pi_{\mu_1,[0,10]} = True$ therefore $\pi_{\mu_1,[0,10]} \in \Pi_{\mu_{act}}$. The activated CBF will progress the system towards the accepting state $s_1$. The system will not reach $s_1$ until $h_1(\textbf{x}_t)$, the predicate function associated with $\pi_{\mu_1,[0,10]}$, becomes non-negative.

\subsection{Control synthesis}
\label{subsec:control}
\noindent \textbf{Finding Control input $\textbf{u}$ (Line \ref{line:commands} of Algo. \ref{algo:synth}):} Given the set of propositions $\Pi_{\mu_{act}}$, we activate CBFs that are associated with those propositions for each robot $i$. Using the time at which a CBF is activated and the position of the robots at time $t$, we activate each pre-constructed barrier function (Section \ref{subsec:beforeExecution}) corresponding to $\Pi_{\mu_{act}}$ if $t$ is in the interval $[t_{int}+a, t_{int} + b]$. The optimization problem we solve to find the control for each robot is shown in  eqn. \ref{cost} where $cfb_{\Psi_i}$ is the combination of all activated CBFs for robot $i$ in the system found from eqn. \ref{combine}.  Eqn. \ref{cost} describes the optimization problem where a control law $u_i$ is found that ensures that $cbf_{\Psi_i}(\textbf{x}_t) \geq 0 \ \forall t$. 
\hspace*{-0.5mm}
\begin{equation}
\begin{split}
     & \min\limits_{\textbf{u}_i \in \textbf{U}_i} \parallel \textbf{u}_i -\hat{\textbf{u}}_i \parallel s.t. \\
     \frac{\partial cbf_{\Psi_i}(\textbf{x}_t)^T}{\partial \textbf{x}}f(\textbf{x},&\textbf{u}) + \frac{\partial cbf_{\Psi_i}(\textbf{x}_t)}{\partial t} \geq -\nu(cbf_{\Psi_i}(\textbf{x}_t))
\label{cost}
\end{split}
\end{equation}
where the nominal controller for each robot $\hat{\textbf{u}}_i$ \dg{is determined using the safe-set associated with the activated CBF. We find a goal point within the safe-set and choose $\hat{\textbf{u}}_i$ in the direction of the goal point with a magnitude equal to the maximum control input for the robot.} If the optimization problem is not feasible, it means we cannot find a control input that satisfies the specification and we stop the execution and provide feedback to the user. 

\begin{figure}[h]
  \begin{subfigure}[b]{0.48\columnwidth}
    \includegraphics[width=\textwidth]{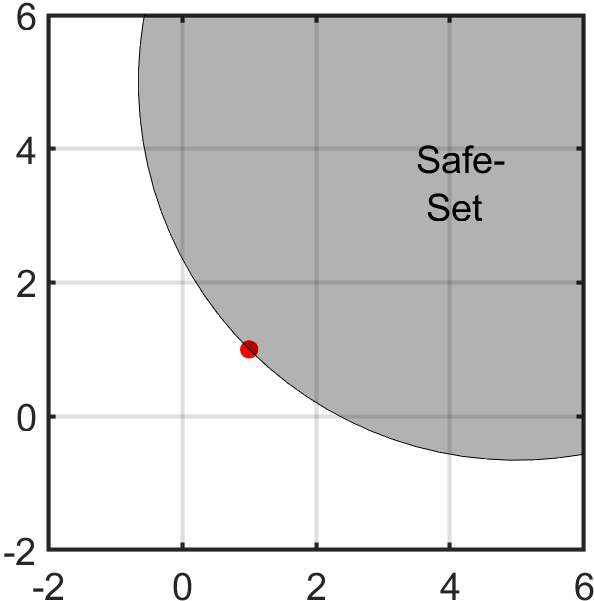}
    \caption{Initial position of the robot (filled red circle) at $t_0$ and corresponding safe set}
    \label{fig:1}
  \end{subfigure}
  ~
  \begin{subfigure}[b]{0.48\columnwidth}
    \includegraphics[width=\textwidth]{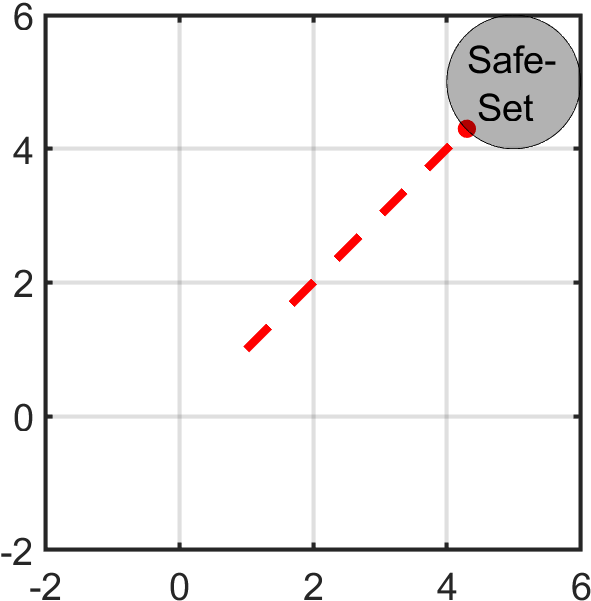}
    \caption{Trajectory from $t_0$ to $t_{final}$ after $alarm$ is sensed and the corresponding safe set at $t_{final}$}
    \label{fig:2}
  \end{subfigure}
  \caption{Safe sets associated with the CBF and trajectory of the robot at the time when the robot senses $alarm$ ($t_0$) and at $t_{final}$ for the single-robot example}
  \label{barrier}
\end{figure}

Figure \ref{barrier} shows the trajectory of an execution of the motivating example. The safe-set associated with the CBF is represented as a circle at time $t_0$ when the proposition $alarm$ becomes $True$, and at $t_{final}=t_0+10$. 


\begin{figure*}
\centering
\begin{subfigure}[t]{0.66\columnwidth}
    \centering
    \includegraphics[width=\textwidth]{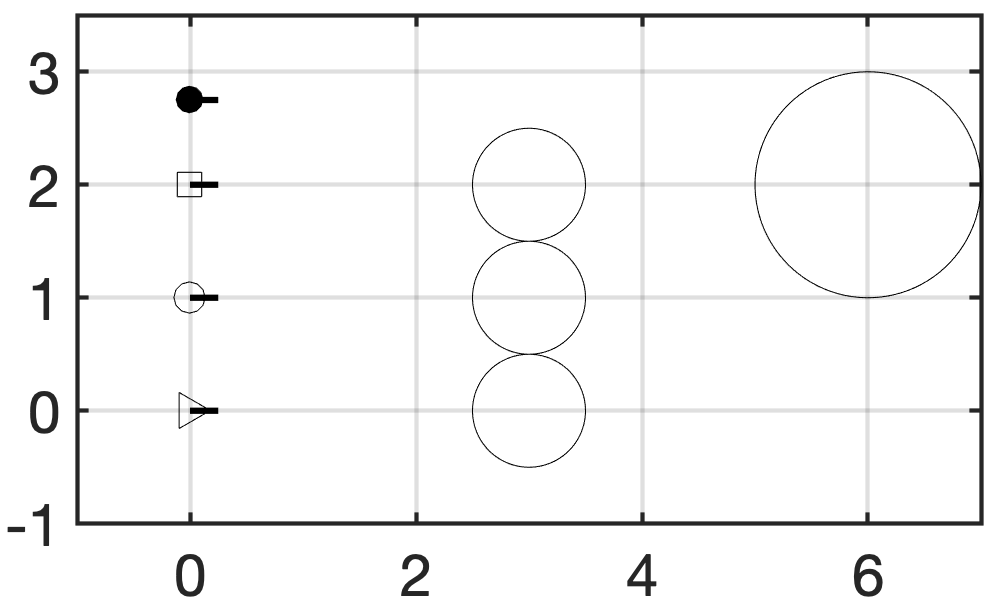}
    \caption{t = 0}
    \label{sim1a}
\end{subfigure}
~ 
\begin{subfigure}[t]{0.66\columnwidth}
    \centering
    \includegraphics[width=\textwidth]{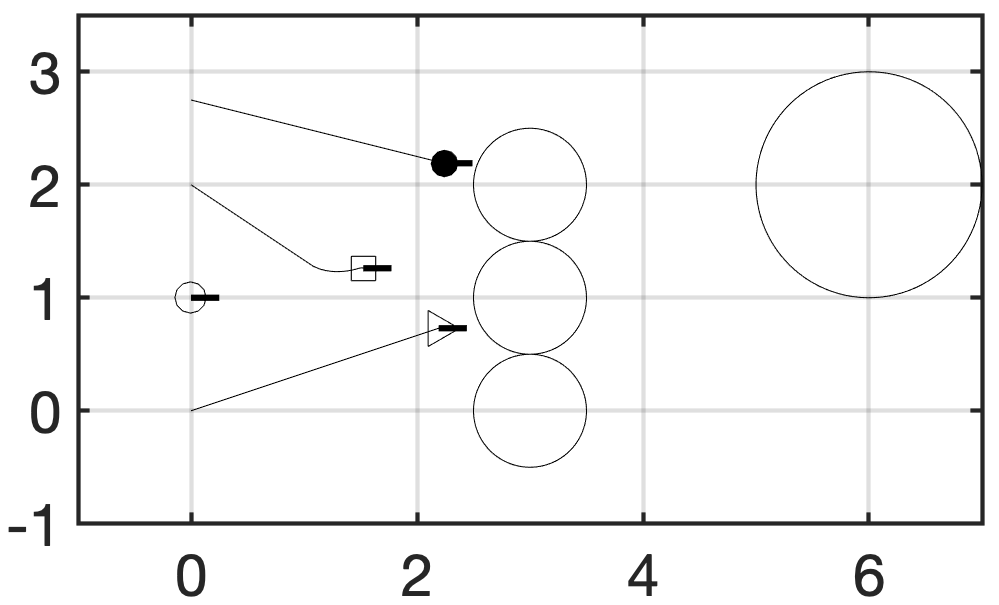}
    \caption{t = 3.5}
    \label{sim1b}
\end{subfigure}
~ 
\begin{subfigure}[t]{0.66\columnwidth}
    \centering
    \includegraphics[width=\textwidth]{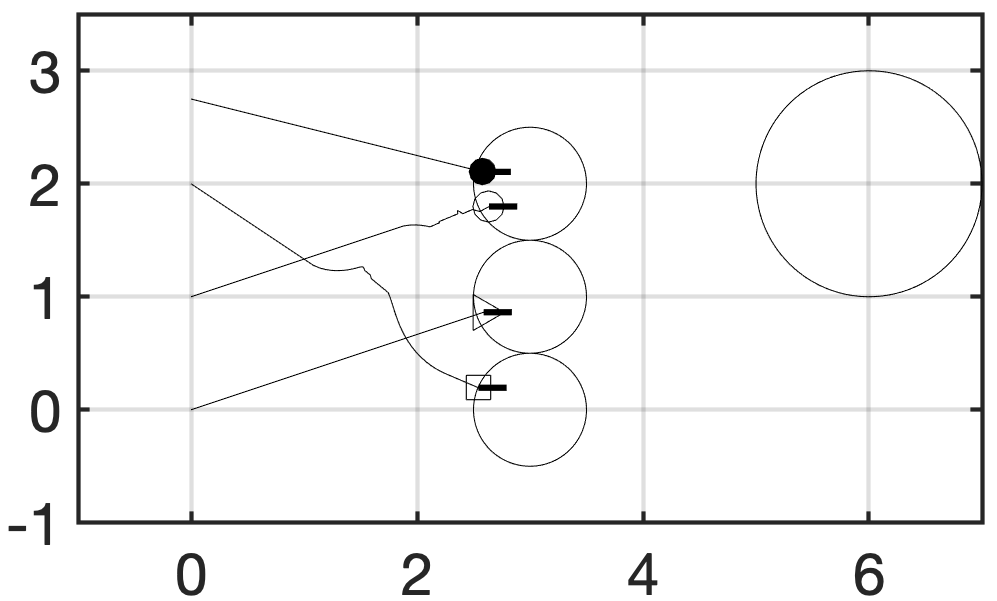}
    \caption{t = 10}
    \label{sim1c}
\end{subfigure}

\hfill 

\begin{subfigure}[t]{0.66\columnwidth}
    \centering
    \includegraphics[width=\textwidth]{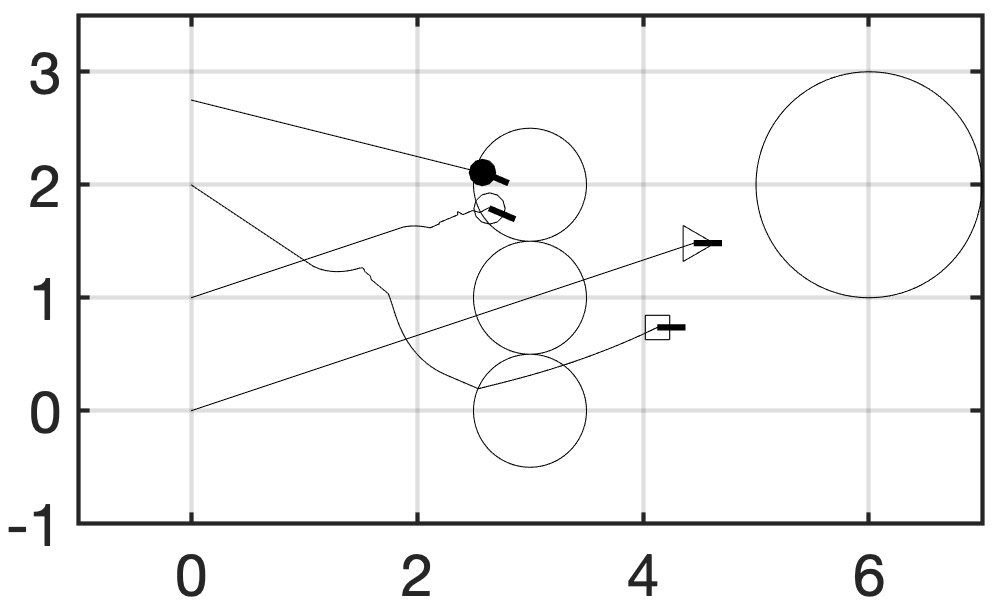}
    \caption{t = 13}
    \label{sim1d}
\end{subfigure}
~ 
\begin{subfigure}[t]{0.66\columnwidth}
    \centering
    \includegraphics[width=\textwidth]{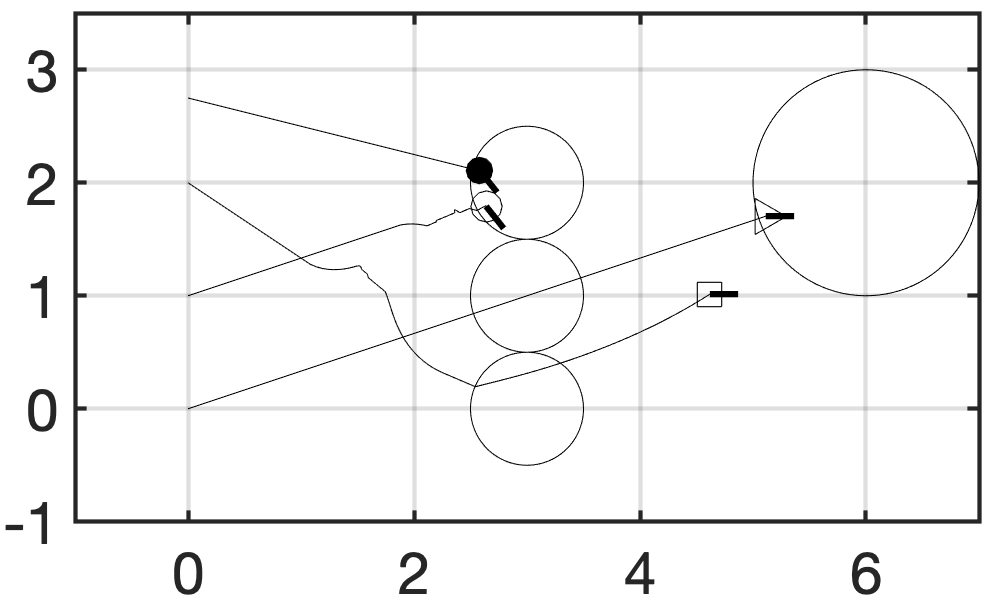}
    \caption{t = 16}
    \label{sim1e}
\end{subfigure}
~ 
\begin{subfigure}[t]{0.66\columnwidth}
    \centering
    \includegraphics[width=\textwidth]{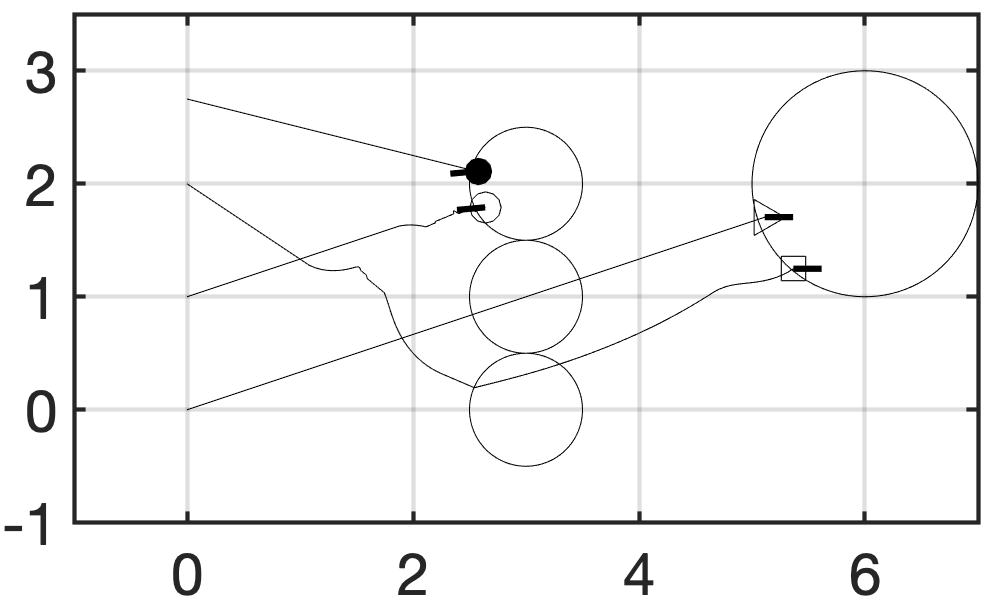}
    \caption{t = 30}
    \label{sim1f}
\end{subfigure}

\caption{Figure \ref{sim1a} shows the initial position of robot 1 (triangle), robot 2 (unfilled circle), robot 3 (square), and robot 4 (filled circle). Figure \ref{sim1b} shows robots 1,3, and 4 progressing towards satisfying $\Psi_1, \Psi_3$, and $\Psi_4$. Robot 3 has to change its path to avoid colliding with robot 1. In figure \ref{sim1c} the robots have satisfied $\Psi_1,..., \Psi_4$. At $t \approx 12$ the robots sense $approach$ and robots 1 and 3 begin to satisfy $\Psi_{approach_1}$ and $\Psi_{approach_3}$. At $t \approx 14$ the robots sense $align$ and robots 2 and 4 begin to satisfy $\Psi_{align_2}$ and $\Psi_{align_4}$ (\ref{sim1d} and \ref{sim1e}). Figure \ref{sim1f} shows the configuration of the robots at $t = 30$.}
\label{sim1}

\end{figure*}

\section{Feedback and Guarantees}\label{sec:feedback}
We give several forms of feedback regarding the feasibility of satisfying a task given the Event-based STL specification and the properties of the robots such as their dynamics and control bounds. We classify the feedback as \emph{a priori} feedback and run-time feedback.
\par
\textbf{A priori feedback:} 
We \hkg{provide two types of feedback based solely on the specification: (i) possibly conflicting CBFs and (ii) inadmissible environment behaviors.} \hkg{For the possibly conflicting CBF,} we examine the set $\Pi_{\mu_{act}}$ of each transition in $B_{\Psi_{LTL}}$. If, for a given transition, the sets $h_i(\textbf{x}_t)$ associated with $\Pi_{\mu_{act}}$ are non-intersecting, we provide feedback to the user that there might not be a control input that satisfies all the associated CBFs. This feedback is conservative as it does not take into account the timing of the STL formulas; depending on the timing, the task may or may not be feasible. 

\dg{\hkg{We place no assumptions on the timing of the environment events; however, depending on the specifications, there might be event timings that cause the LTL formula to be unsatisfiable. To alert the user to such situations,} for each state, we examine the possible transitions out of the state and find \hkg{combinations of} environment events that are not allowed \hkg{i.e., none of the transitions out of the state allows such a combination. We provide these combinations to the user}.}
\par
\textbf{Run-time feedback:} Unknown disturbances such as other robots in the system, environment disturbances, or deadlocks can prevent the system from completing the task. We provide feedback on the feasibility of satisfying a task during an execution given the configuration of the system, the timing requirements, and the control bounds of the robots. 

At each time-step, we calculate how far the system is from satisfying the predicate functions by evaluating $h(\textbf{x}_t)$. We then compare this distance to the largest distance the system can move in state space given the bounds on the control and the time remaining to satisfy the predicate $$\parallel u_{max}\parallel(b + t_{int} -t)$$ If the distance to the predicate function is larger than the maximum distance the system can travel, it means the system will fail the task and we provide feedback to the user. \dg{In this situation a solution to the optimization problem may still be found even though the specification will be violated at a future time. While for the CBF template we present in this paper this in not the case, other templates may define larger safe-sets that shrink faster, leading to the robot not being able to remain in the safe-set in the future.}  

We check the distance from each individual predicate; however, even if all predicates are within reach,  when combining several CBFs in the optimization problem eqn. \ref{cost}, it may become infeasible. This might happen when the system is trying to reach two predicates that require motion in opposite direction. In these cases, we stop the system and provide feedback to the user. 

\hkg{\textbf{Completeness}: Our specification formalism and control synthesis approach enable a user to capture complex tasks and automatically execute them; however, our approach is not complete. There may be a control that could satisfy the task in situations in which we have determined that we cannot find a control input. This can occur, for example, when there are conflicting predicates that may be achieved by adding a prioritization scheme. The feedback we are providing before and during execution is meant to mitigate possible execution failures.}

\section{Simulation Results}
\label{sec:sim}

\subsection{Simulation Example Description}
We consider 4 holonomic robots that operate in a shared environment. They are performing the multi-robot 
task described in Sec \ref{sec:problem}. The robots do not collaborate and each robot only has information about the position of the other robots. The initial state $\textbf{x}_0 = [x_1,y_1,\theta_1, ... ,x_4,y_4,\theta_4]$ of the system is $\textbf{x}_0 = [0,0,0,0,1,0,0,2,0,0,2.75,0]$ and the velocity bound $\textbf{U} = \pm[u_{x1},u_{y1},u_{\theta 1},...,u_{x4},u_{y4},u_{\theta 4}]$ of the system is $\textbf{U}= \pm[0.7,0.7,0.5,0.9,0.9,0.5,0.65,0.65,0.5,0.8,0.8,0.5]$. The B\"{u}chi automaton took 1:15 minutes to compute on a 2.3 GHz Quad-Core CPU with 8 GB of RAM and contains 281 states, 21,121 transitions, and 14 CBFs. 

\subsection{Simulation Results}
Figure \ref{sim1} shows the trajectory of the system at different time steps. All robots are able to satisfy their individual tasks while avoiding collisions as defined in $\Psi_{collision_{ij}}$. The robots were able to proceed to their goal regions represented by the circular regions without collision. This simulation was run at 10Hz and the controllers for all robots in the system took approximately 0.07 seconds to compute. The simulation was run on a 2.3 GHz Quad-Core CPU with 8 GB of RAM. \dg{\hkg{The }a priori feedback is that if $approach_1$ or $approach_3$ are sensed before \hkg{the predicates in} $\Psi_1$ or $\Psi_3$ are satisfied, the specification may be violated.}

\section{Physical Demonstration}
\label{sec:phys}
\subsection{Example Description}
To further show the expressive power of Event-based STL and the feedback we can generate we conduct a physical demonstration with two iRobot Creates. We consider the following Event-based STL specification for the multi-robot system

\begin{itemize}
    \item $\Psi_1 = F_{[0,15]}(\parallel \textbf{x}_1 - [-2,1]^T \parallel < 0.5)$
    \item $\Psi_2 = F_{[1,16]}(\parallel \textbf{x}_2 - [2,1]^T \parallel < 0.5)$
    \item $\Psi_3 = G(alarm \Rightarrow F_{[0,10]} (\parallel \textbf{x}_1 - [0,-1]^T \parallel < 0.5))$
    \item $\Psi_4 = G_{[0,25]}(\parallel \textbf{x}_1 - \textbf{x}_2 \parallel > 0.5)$
\end{itemize}

The task is defined as the conjunction of all of the Event-based STL formulas $\Psi = \Psi_1 \wedge \Psi_2 \wedge \Psi_3 \wedge \Psi_4$

\subsection{A priori feedback}
Before executing a run, 
we provide feedback to the user on the feasibility of a task. To do this we check if conflicting CBFs exist that may be activated at the same time during an execution, as outlined in Sec. \ref{sec:feedback}. For the physical demonstration there are several transitions in the B\"{u}chi automaton that activate conflicting CBFs. These conflicting CBFs come from the predicate functions associated with $\Psi_1$ and $\Psi_3$ which can not be satisfied at the same time. This only occurs when the robot senses $alarm$ and $\Psi_1$ has not been satisfied. We alert the user of this potential issue so that they can change the specification accordingly. 

\subsection{Physical Demonstration Results}
\begin{figure}[h!]
\setlength{\textfloatsep}{0.2\baselineskip plus 0.2\baselineskip minus 0.5\baselineskip}
\centering
\begin{subfigure}[t]{0.45\columnwidth}
    \centering
    \includegraphics[width=\textwidth]{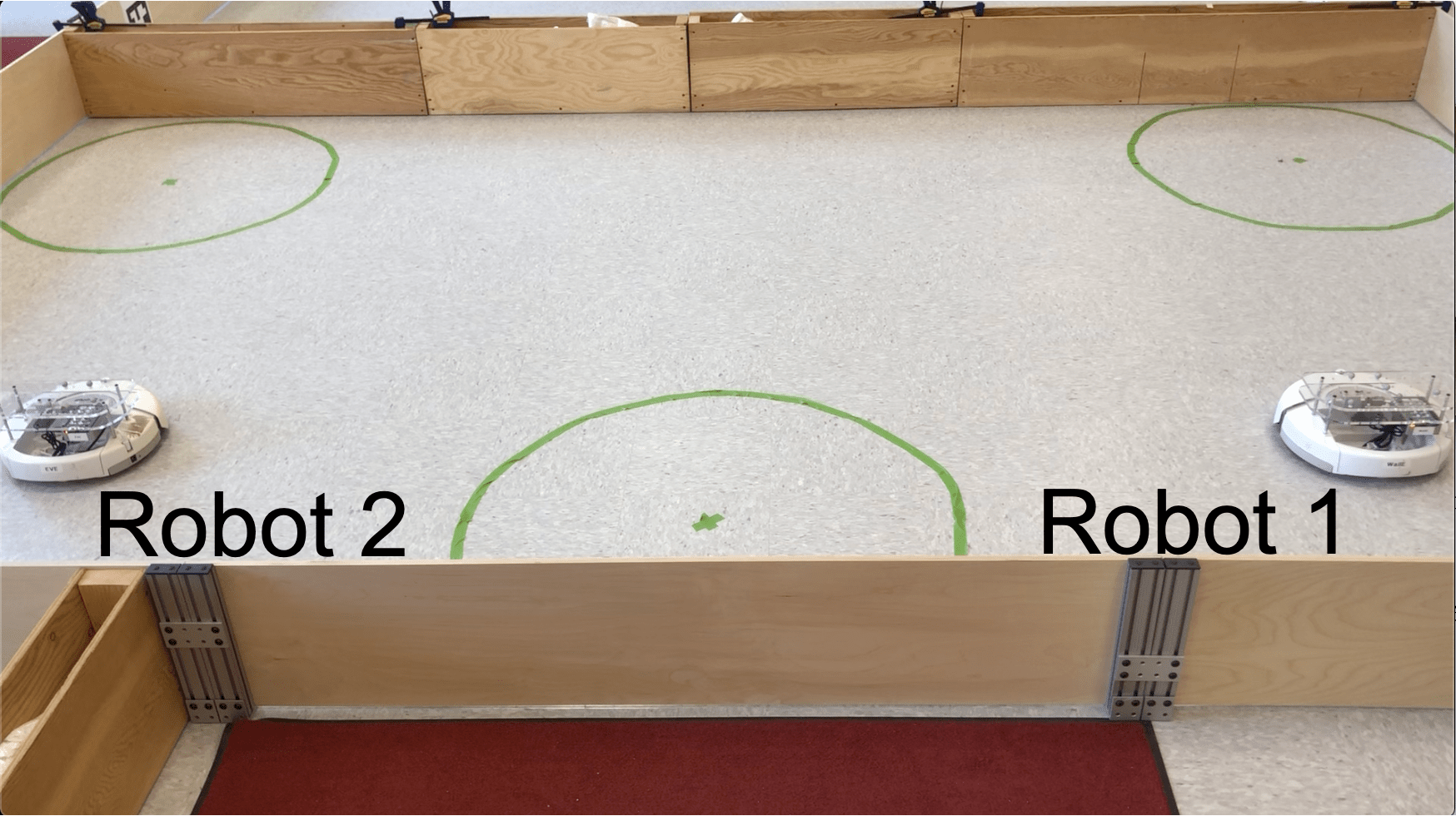}
    \caption{Initial configuration of robot 1 and robot 2}
    \label{phys1a}
\end{subfigure}
~ 
\begin{subfigure}[t]{0.45\columnwidth}
    \centering
    \includegraphics[width=\textwidth]{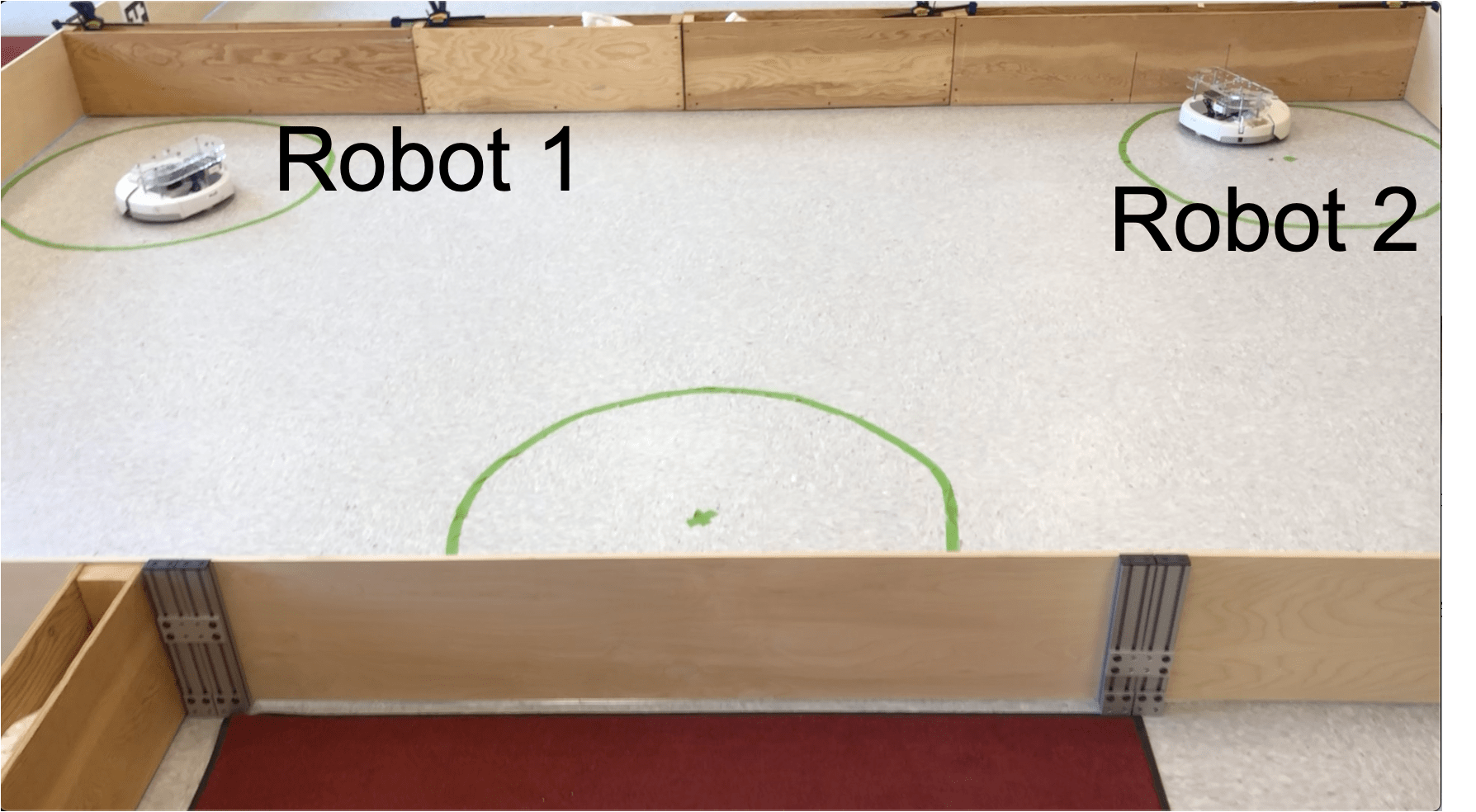}
    \caption{Configuration of the robots at $t = 25$. 
    }
    \label{phys1b}
\end{subfigure}

\caption{The robots do not sense $alarm$ and the robots remain in an accepting states in the safe-sets defined by $\Psi_1$ and $\Psi_2$}
\label{phys1}
\vspace{-15pt}
\end{figure}

The following section describes the results of the physical demonstrations where $alarm$ becomes $True$ at different times. In the first execution $alarm$ never becomes $True$ and the system remains in an accepting state. Snapshots of this run are shown in figure \ref{phys1}.

\begin{figure}[h!]

\centering
\begin{subfigure}[t]{0.45\columnwidth}
    \centering
    \includegraphics[width=\textwidth]{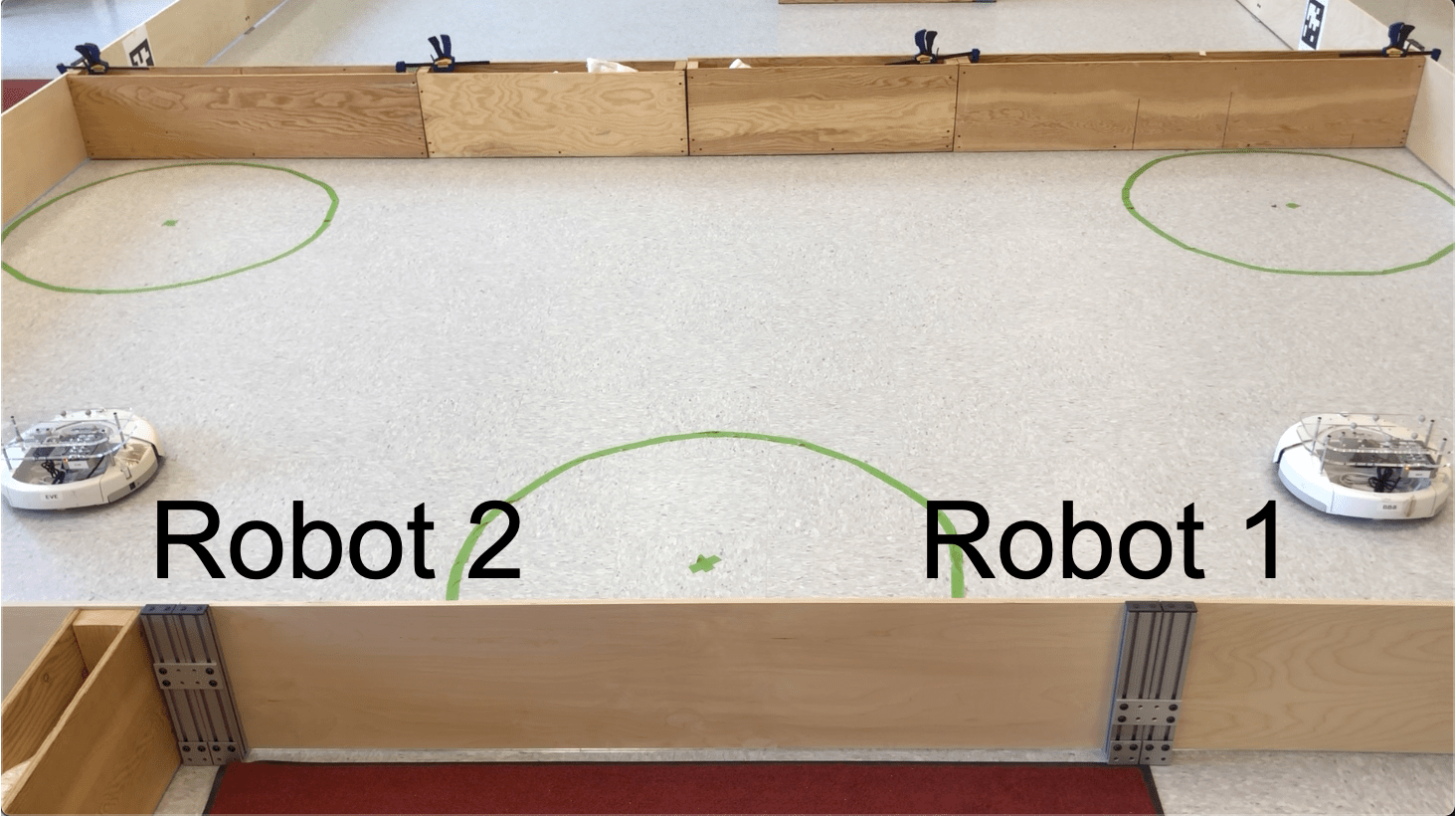}
    \caption{Initial configuration of robot 1 and robot 2}
    \label{phys2a}
\end{subfigure}
~ 
\begin{subfigure}[t]{0.45\columnwidth}
    \centering
    \includegraphics[width=\textwidth]{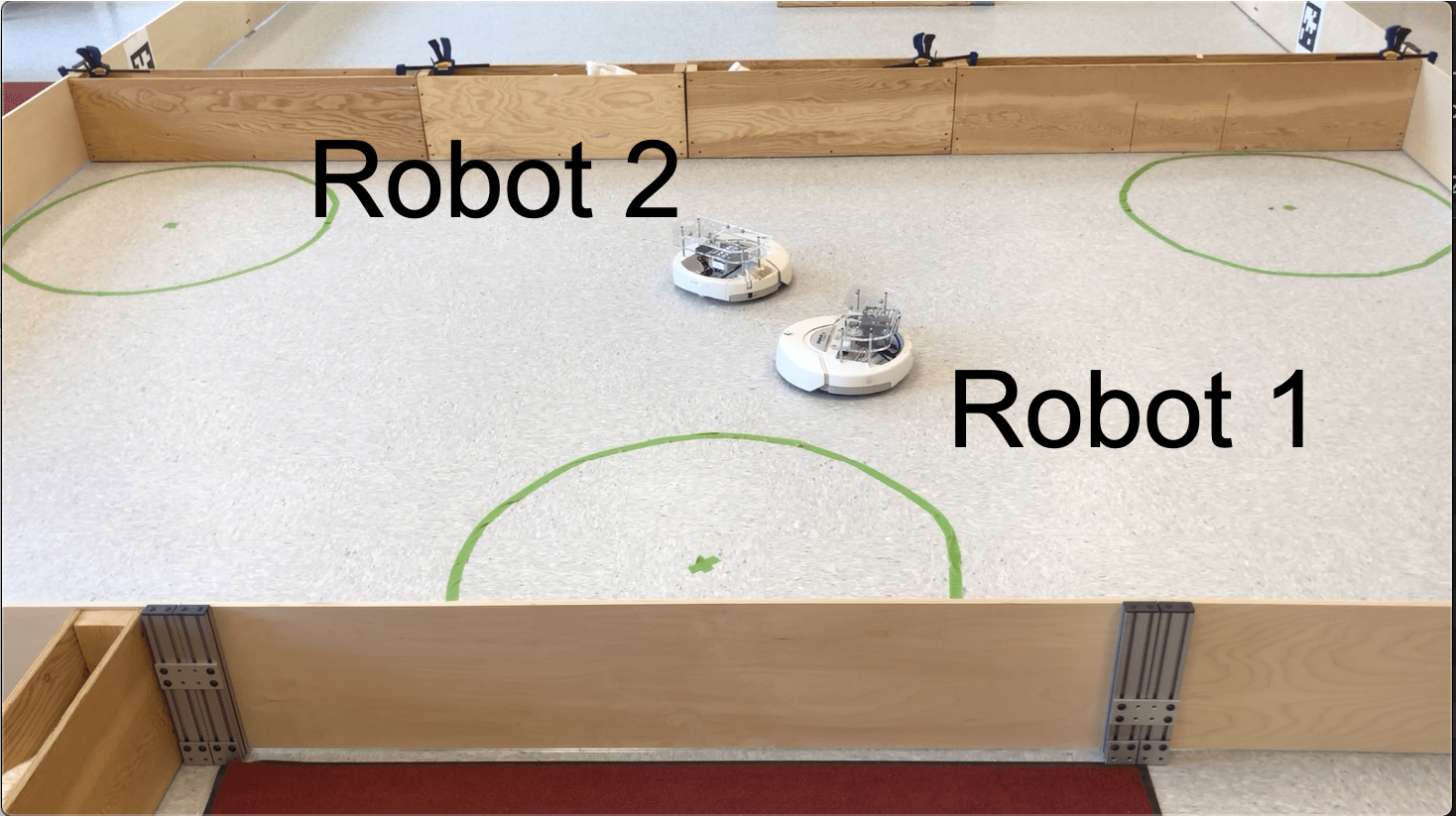}
    \caption{Both robots have to change directions to avoid colliding with each other}
    \label{phys2b}
\end{subfigure}
\hfill 
\begin{subfigure}[t]{0.45\columnwidth}
    \centering
    \includegraphics[width=\textwidth]{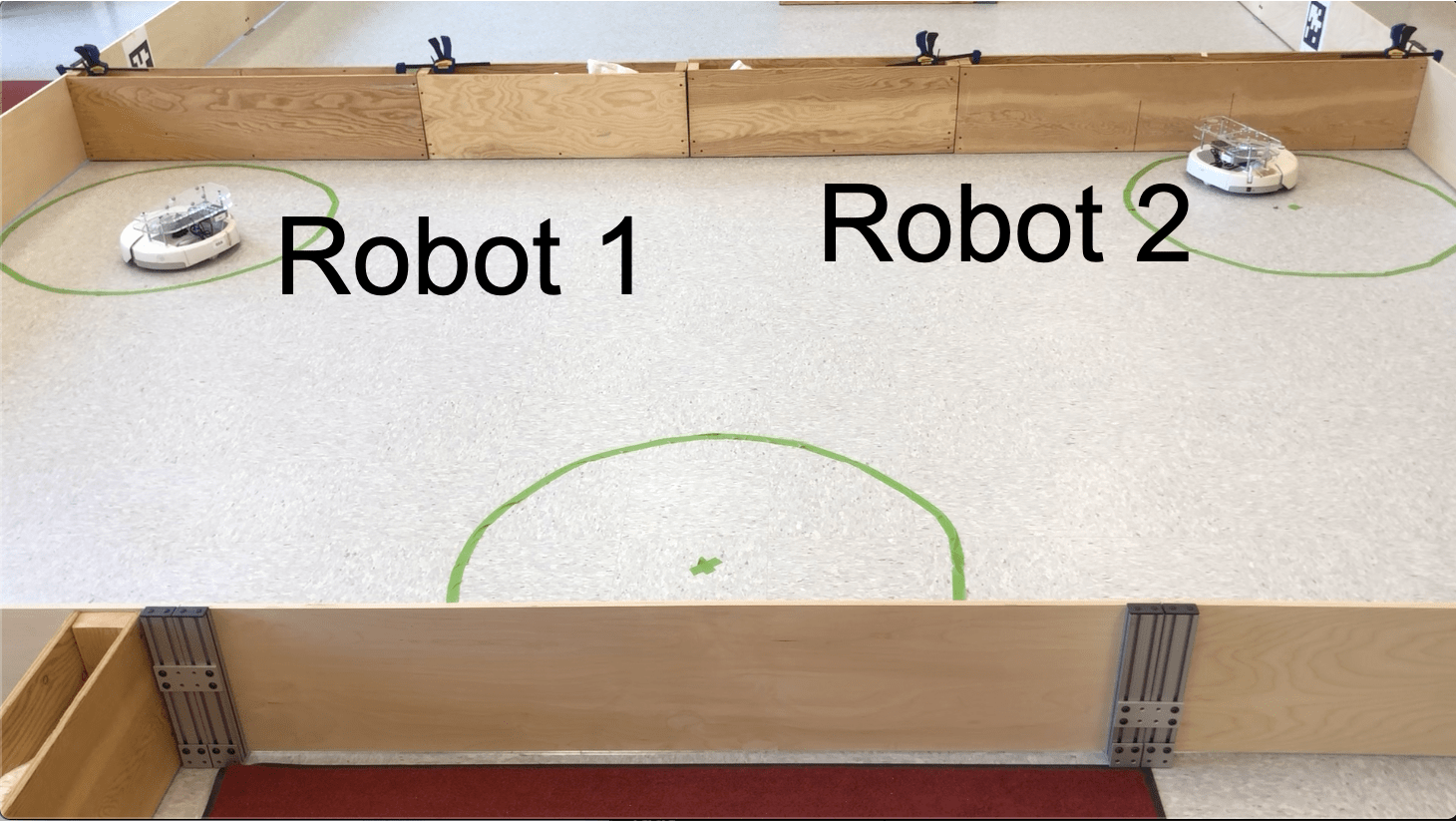}
    \caption{The robots sense $alarm$ after robot 1 satisfies $\Psi_{1}$ at $\approx$ 17}
    \label{phys2c}
\end{subfigure}
~ 
\begin{subfigure}[t]{0.45\columnwidth}
    \centering
    \includegraphics[width=\textwidth]{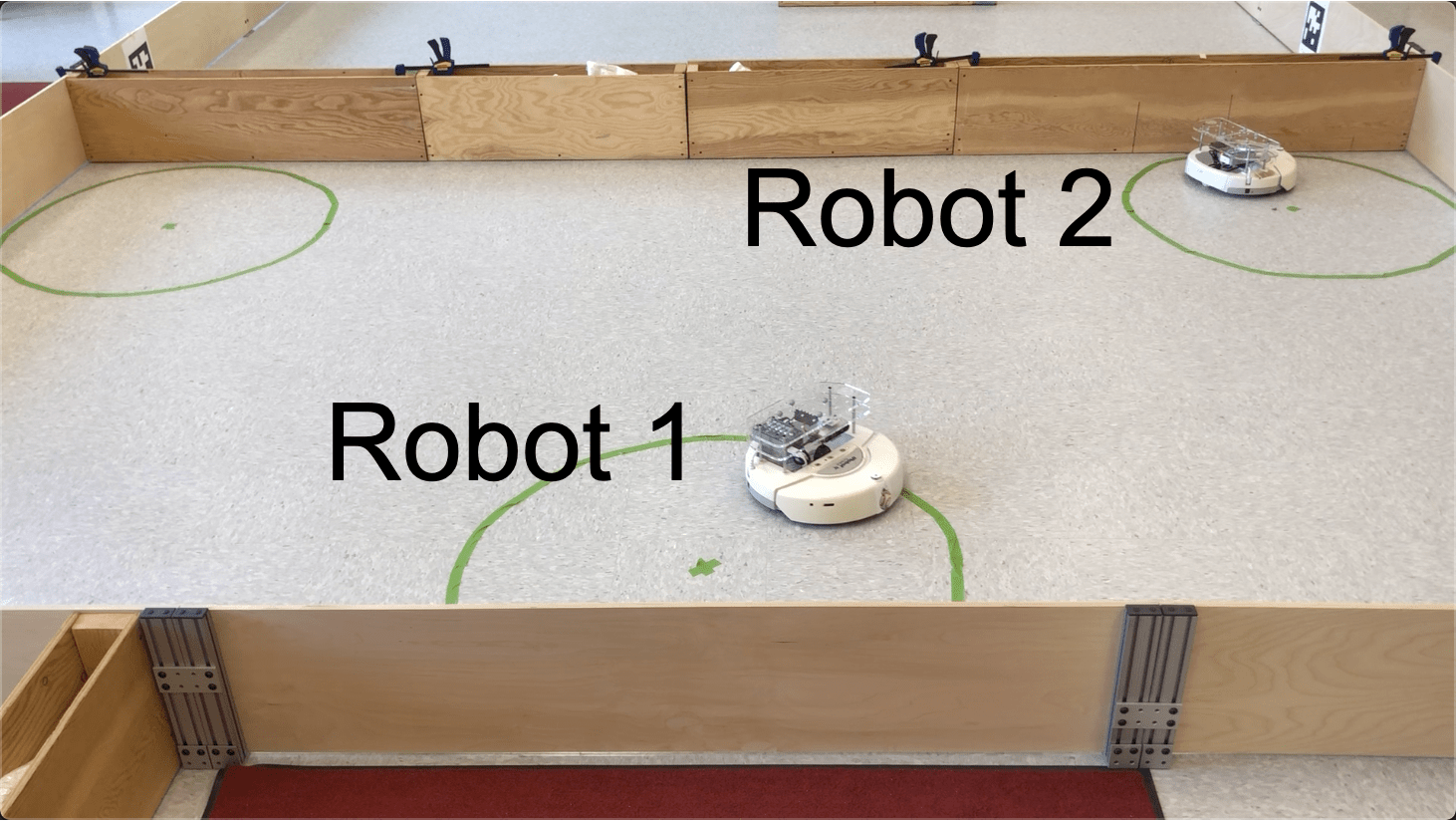}
    \caption{Robots in safe-set satisfying full specification. Robot 1 and robot 2}
    \label{phys2d}
\end{subfigure}

\caption{snapshots of an execution when the robots sense $alarm$ }
\vspace{-15pt}
\label{phys2}
\end{figure}

In the second run the robot senses the $alarm$ event at $t \approx 17$. This is after robot 1 has satisfied $\Psi_1$. Figure \ref{phys2} shows the position of the robots at various timesteps. In this execution the collision avoidance described by $\Psi_4$ can be seen as both robots change their paths so that they do not collide with each other. \dg{During execution there is no run-time feedback given because the robots are always able to satisfy their tasks based on their bounded control inputs.}

\section{Conclusions}
We provide a framework for expressing and synthesizing control for high-level specifications that include reactions to uncontrolled events and bounds on time and control input. To do this we create a specification formalism called Event-based STL and show its capabilities through simulation and physical demonstrations. Because there are bounded control inputs and a possibility of unknown disturbances and environment inputs, we cannot provide a-priori guarantees that a specification can be satisfied. Instead we provide feedback to the user as to why the specification can not be satisfied, when we detect a problem. In future work we will consider specifications in complex environments and work to expand the feedback given to users regarding infeasible tasks and provide suggestions of changes to make the specification satisfiable. 

\bibliographystyle{ieeetr}
\bibliography{Event-BasedSTL.bib}

\end{document}